\documentclass[conference]{IEEEtran}
\IEEEoverridecommandlockouts

\usepackage[utf8]{inputenc}
\usepackage[english]{babel}
\usepackage{enumitem} 
\usepackage{amsthm}
{
\theoremstyle{plain}

}
\usepackage[linesnumbered,ruled,vlined]{algorithm2e}
\usepackage{algpseudocode}
\algrenewcommand\algorithmicrequire{\textbf{Input:}}
\algrenewcommand\algorithmicensure{\textbf{Output:}}
\SetKwRepeat{Do}{do}{while}
\usepackage{cite}
\usepackage{threeparttable}
\usepackage{makecell}
\usepackage{amsmath,amssymb,amsfonts}
\usepackage{amsmath}
\usepackage{graphicx}
\usepackage{bbm}
\usepackage{caption}
\usepackage[outdir=./]{epstopdf}
\usepackage{textcomp}
\usepackage[table,usenames,dvipsnames]{xcolor}

\usepackage{mathtools}

\usepackage[top=1in,bottom=0.75in,right=0.75in,left=0.75in]{geometry}
\usepackage{subfigure}
\usepackage{authblk}
\renewcommand{\theenumi}{\alph{enumi}}
\SetKwInput{KwInput}{Input}                
\SetKwInput{KwOutput}{Output}              
\def\BibTeX{{\rm B\kern-.05em{\sc i\kern-.025em b}\kern-.08em
    T\kern-.1667em\lower.7ex\hbox{E}\kern-.125emX}}

\usepackage{balance}

\usepackage[breaklinks=true, colorlinks, bookmarks=true, citecolor=Black, urlcolor=Violet,linkcolor=Black]{hyperref}

\newcommand{\prl}[1]{\left(#1\right)}

\newcommand{\crl}[1]{\left\{#1\right\}}

\newcommand{\calA}{{\cal A}}
\newcommand{\calB}{{\cal B}}

\newcommand{\calD}{{\cal D}}

\newcommand{\calI}{{\cal I}}

\newcommand{\calS}{{\cal S}}

\newcommand{\bfa}{\mathbf{a}}

\newcommand{\bfr}{\mathbf{r}}
\newcommand{\bfs}{\mathbf{s}}

\newcommand{\bfy}{\mathbf{y}}

\newcommand{\bftheta}{\boldsymbol{\theta}}

\newcommand{\bfpi}{\boldsymbol{\pi}}

\newcommand{\bfC}{\mathbf{C}}

\newcommand{\bbR}{\mathbb{R}}

\begin{document}

\title{Coding for Distributed Multi-Agent Reinforcement Learning%
\thanks{This work is supported by the National Science Foundation (NSF) under grants 1953048 and 1953049, the San Diego State University under the University Grants Program, and ARL DCIST CRA W911NF-17-2-0181.}
}

\author{Baoqian Wang$^{1}$ \and Junfei Xie$^{2}$ \and Nikolay Atanasov$^{3}$%
    \thanks{$^{1}$ Baoqian Wang is with the Department of Electrical and Computer Engineering, University of California, San Diego and San Diego State University, La Jolla, CA, 92093 (e-mail: \textit{bawang@ucsd.edu}).}%
    \thanks{$^{2}$ Junfei Xie is with the Department of Electrical and Computer Engineering, San Diego State University, San Diego, CA, 92182 (e-mail: \textit{jxie4@sdsu.edu}).}%
    \thanks{$^{3}$ Nikolay Atanasov is with the Department of Electrical and Computer Engineering, University of California San Diego, La Jolla, CA, 92093 (e-mail: \textit{natanasov@ucsd.edu}).}
}


\maketitle

\begin{abstract}
This paper aims 
to mitigate  straggler effects in synchronous distributed learning for multi-agent reinforcement learning (MARL) problems. Stragglers arise frequently in a distributed learning system, due to the existence of various system disturbances such as slow-downs or failures of compute nodes and communication bottlenecks. To resolve this issue, we propose a coded distributed learning framework, which speeds up the training of MARL algorithms in the presence of stragglers, while maintaining the same accuracy as the centralized approach. As an illustration, a coded distributed version of the multi-agent deep deterministic policy gradient (MADDPG) algorithm is developed and evaluated. Different
coding schemes, including maximum distance separable (MDS) code, random sparse code, replication-based code, and regular low density parity check (LDPC) code are also investigated. 
Simulations in several multi-robot problems demonstrate the promising performance of the proposed framework.  
\end{abstract}


\section{Introduction}
\label{sec:introduction}

Many real-life applications involve interaction among multiple intelligent systems, such as collaborative robot teams \cite{bonnet2017multi}, internet-of-things devices \cite{forestiero2017multi}, agents in cooperative or competitive games \cite{lowe2017multi}, and traffic management devices \cite{chu2019multi}. Reinforcement learning (RL) \cite{RLBook} is an effective tool to optimize the behavior of intelligent agents in such applications based on reward signals from interaction with the environment. Traditional RL algorithms, such as Q-Learning \cite{tan1993multi} and policy gradient \cite{lowe2017multi}, can be scaled to multiple agents by simultaneous application to each individual agent. However, learning independently for each agent performs poorly as the environment is non-stationary from the perspective of a single agent due to the actions of the other agents \cite{lowe2017multi,matignon2012independent}. Multi-agent reinforcement learning (MARL) \cite{tan1993multi} focuses on mitigating these challenges by adding other agents' policy parameters to the Q function \cite{tesauro2004extending} or relying on importance sampling \cite{foerster2017stabilising}. Yang et al. \cite{yang2018mean} propose a mean-field Q learning algorithm, which uses Q functions defined only with respect to an agent's own state and those of its neighbors instead of all agent states. The multi-agent deep deterministic policy gradient (MADDPG) \cite{lowe2017multi} is an extension of the deep deterministic policy gradient (DDPG) algorithm \cite{ddpg} to a multi-agent setting. MADDPG uses a Q function that depends on all agent states and actions but local control policies, defined over the state and action of an individual agent. One key challenge faced by centralized MARL approaches is that the training computational complexity scales with the number of agents in the environment.

Distributing the training workload is a promising solution for accelerating RL and MARL training that has attracted a lot of attention recently. The first massively distributed architecture for deep RL is presented in \cite{nair2015massively}. It relies on an off-policy deep Q network algorithm that uses multiple actors and learners running in parallel. The asynchronous advantage actor-critic (A3C) algorithm \cite{mnih2016asynchronous} allows decorrelating the training data from multiple environments executed asynchronously and thus can be used for either on-policy or off-policy learning. Multiple variants of A3C have been developed, including GPU A3C (GA3C) \cite{babaeizadeh2016reinforcement}, that allows A3C to be trained on hybrid CPU-GPU architecture, and advantage actor-critic (A2C) \cite{a2c}, which achieves decorrelation with synchronous training. Distributed learning has received much less attention in the MARL setting. Sim{\~o}es et al. \cite{simoes2020multi} extend the A3C framework to multi-agent systems by running multiple compute nodes in parallel to update the parameters asynchronously, with each node simulating multiple agents interacting with the environment. In \cite{chu2019multi}, a multi-agent version of A2C with synchronous learning is investigated to address a traffic signal control problem.

In the aforementioned distributed MARL algorithms, learning is performed either synchronously or asynchronously. A major challenge for synchronous learning is its vulnerability to system disturbances, such as slow-downs or failures of individual learner nodes and communication bottlenecks in the network traffic. Learners that suffer from such issues become \emph{stragglers} that can significantly delay the learning process. Asynchronous learning \cite{ho2013more, li2014communication} can help mitigate the impact of stragglers but suffers from other limitations, including slower convergence rate, lower accuracy, and more challenging analysis and debugging \cite{a2c,tan1993multi}. 

We propose \emph{coding theory} strategies \cite{lee16speeding,lee17high} to mitigate straggler effects in synchronous distributed learning for MARL problems. Coding techniques have been successful in improving the resilience of communication, storage, and cache systems to uncertain system disturbances \cite{li15coded, liu2020esetstore, zhu2013speedup}. Such techniques have recently been applied to speed up distributed computation tasks in the presence of stragglers, including matrix-vector \cite{lee16speeding,lee18speeding} and matrix-matrix \cite{lee17high} multiplication, linear inverse problems \cite{yang17coded}, multivariate polynomials \cite{yu2019lagrange, rudow2020locality}, convolution \cite{dutta2017coded}, and gradient descent \cite{tandon17gradient, li2018near, bitar2020stochastic}. Nevertheless, the merits of coding techniques remain to be explored in the RL and MARL settings.

The main \emph{contribution} of this work is a coded distributed learning framework that can be applied with any policy gradient method to solve MARL problems efficiently despite possible straggler effects. As an illustration, we apply the proposed framework to create a coded distributed version of MADDPG \cite{lowe2017multi}, a state-of-the-art MARL algorithm. Furthermore, to gain a comprehensive understanding of the benefits of coding in distributed MARL, we investigate various codes, including the maximum distance separable (MDS) code, random sparse code, replication-based code, and regular low density parity check (LDPC) code. Simulations in several multi-robot problems, including cooperative navigation, predator-prey, physical deception and keep away tasks \cite{lowe2017multi}, indicate that the proposed framework speeds up the training of policy gradient algorithms in the presence of stragglers, while maintaining the same accuracy as a centralized approach.

\section{Multi-Agent Reinforcement Learning}
\label{sec:problem}

This section introduces the MARL problem and reviews the class of policy gradient algorithms for MARL problems.


\subsection{Problem Formulation}

In MARL, multiple agents learn to achieve specific goals by interacting with the environment. Let $M$ be the number of agents. Denote the state and action of agent $i \in [M] := \{1,\ldots,M\}$ at time $t$ by $s_{i,t}\in \mathcal{S}_i$ and $a_{i,t} \in \calA_i$, respectively, where $\calS_i$ and $\calA_i$ are the corresponding state and action spaces. Let $\bfs_t := (s_{1,t},\ldots, s_{M,t}) \in \mathcal{S} := \prod_{i \in [M]} \mathcal{S}_{i}$ and $\bfa_t := (a_{1,t}, \ldots, a_{M,t}) \in \mathcal{A} := \prod_{i \in [M]} \mathcal{A}_{i}$ denote the joint state and action of all agents. At any time $t$, a joint action $\bfa_t$ applied at state $\bfs_t$ triggers a transition to a new state $\bfs_{t+1} \in \calS$ according to an (unknown) conditional probability density function (pdf) $p(\bfs_{t+1} | \bfs_t, \bfa_t)$. After each transition, each agent $i$ receives a reward $r_i(\bfs_t,\bfa_t) \in \bbR$, where the reward functions $r_i$ may be different for different agents.

Given a joint state $\bfs$, the objective of each agent $i$ is to choose a stochastic policy, specified by a pdf $\pi_i(a_i|\bfs)$ over the agent's action space $\calA_i$, such that the expected cumulative discounted reward:
\begin{equation}
\label{eq:value_function}
V^{\bfpi}_i(\bfs) := \mathbb{E}_{\substack{\bfs_t\sim p\\ \bfa_t\sim \bfpi}} \biggl[\sum_{t=0}^{\infty}\gamma^t r_i(\bfs_t,\bfa_t) |\bfs_0=\bfs\biggr]
\end{equation}
is maximized. In \eqref{eq:value_function}, $\gamma \in (0,1]$ is a discount factor,  $\bfpi := \left(\pi_{1}, \ldots, \pi_{M}\right)$ denotes the joint policy of all agents. The function $V^{\bfpi}_i(\bfs)$ is known as the value function of agent $i$ associated with joint policy $\bfpi$. Alternatively, an optimal policy $\pi_i^*$ for agent $i$ can be obtained by maximizing the action-value function:
\begin{equation}
Q_i^{\bfpi}(\bfs,\bfa) := \mathbb{E}_{\substack{\bfs_t\sim p\\ \bfa_t\sim \bfpi}}\biggl[\sum_{t=0}^{\infty}\gamma^t r_i(\bfs_t,\bfa_t) |\bfs_0=\bfs, \bfa_0 = \bfa\biggr] \nonumber
\end{equation}
and setting $\pi_i^*(a_i | \bfs) \in  \arg\max_{a_i} \max_{\bfa_{-i}} Q_i^{*}(\bfs,\bfa)$, where $Q_i^{*}(\bfs,\bfa) := \max_{\bfpi} Q_i^{\bfpi}(\bfs,\bfa)$ and $\bfa_{-i}$ denotes the actions of all agents except $i$.

\subsection{Policy Gradient Methods}
There are two major classes of MARL algorithms: \emph{value-based} and \emph{policy-based}. Value-based algorithms aim to approximate the optimal action-value function $Q_i^{*}(\bfs,\bfa)$. Examples include extensions of Q-learning \cite{watkins1992q}, DQN \cite{fan2020theoretical}, and SARSA \cite{RLBook} to multi-agent settings, such as Independent Q-learning \cite{tan1993multi}, Inter-Agent Learning \cite{foerster2016learning}, and multi-agent SARSA \cite{wang2014integrating}. Policy-based algorithms directly optimize over the space of policy functions. A general approach, known as \emph{policy gradient} \cite{RLBook}, is to define parametric policy functions $\pi_i(a_i | \bfs ; \theta_i)$, using linear feature or neural network parameterization, and update the parameters along the gradient of the value function: $\theta_i \gets \theta_i + \alpha \nabla_{\theta_i} V_i^{\bfpi}(\bfs)$. The value function gradient is obtained via the policy gradient theorem \cite{RLBook}:
\begin{equation}
    \nabla_{\theta_i} V^{\bfpi}_i(\bfs) =\mathbb{E}_{\substack{\bfs_t\sim p\\\bfa_t\sim \bfpi}}\bigl[\nabla_{\theta_i} \!\prl{\log \pi_{i}(a_{i,t} | \bfs_t; \theta_i)} Q_i^{\bfpi}(\bfs_t, \bfa_t)\bigr]. \nonumber
\end{equation}
Different approaches exist for estimating $Q_i^{\bfpi}(\bfs_t, \bfa_t)$ in the above equation, such as the \emph{REINFORCE} algorithm \cite{REINFORCE}, which uses sample returns  $\sum_{t=0}^T\gamma^t r_i(\bfs_t,\bfa_t)$  from several episodes of length $T$, critic methods \cite{konda2000actor} that parameterize and approximate $Q_i^{\bfpi}(\bfs, \bfa)$ in addition to the policy, or the  trust region policy optimization (TRPO) algorithm \cite{schulman2015trust}, which ensures that a policy improvement condition is satisfied.

\section{Coded Distributed Learning for MARL}
\label{sec:coded_dmarl}
In this section, we introduce a coded distributed learning framework that can be incorporated with any general policy gradient method to solve the aforementioned 
MARL problem efficiently in the presence of stragglers. Before we describe this framework, we first overview a distributed learning framework for MARL without coding.

\subsection{Uncoded Distributed Learning for MARL}
\begin{figure}[t]
 \centering
	\includegraphics[width=0.8\linewidth]{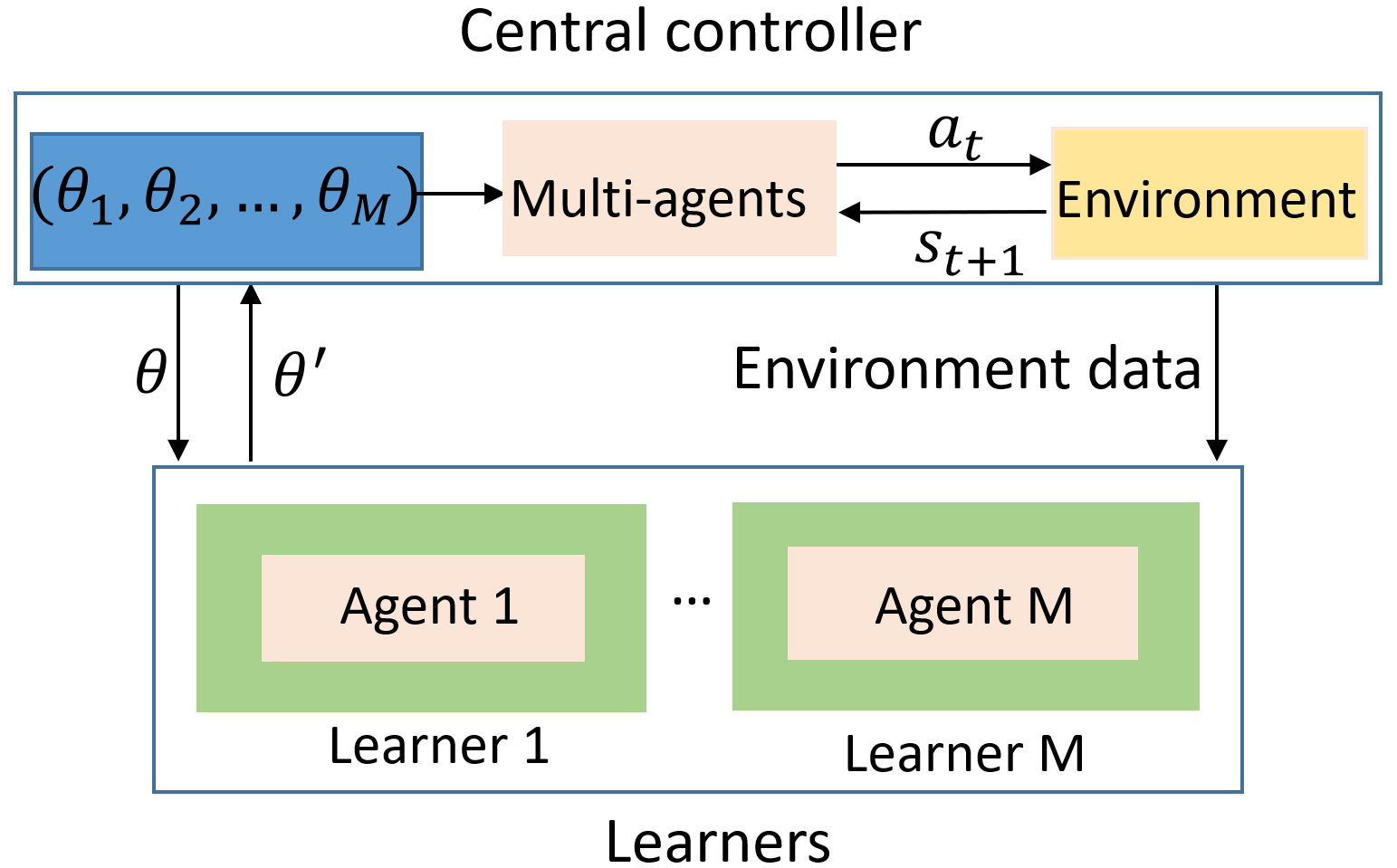} \vspace*{-0.25em}
	\caption{Illustration of uncoded distributed learning for MARL.}
	\label{fig:dis_learning}
\end{figure}

A  distributed learning system (see Fig. \ref{fig:dis_learning} for an illustration) for RL/MARL consists of a \emph{central controller} and multiple \emph{learners}. The central controller is a server that maintains all agent parameters, 
and uses them to construct policies for the agents. 
In each training iteration, it executes multiple episodes and sends both the collected environment data and the parameters of all agents 
to the learners. Once a learner 
receives these data, it 
updates the agent parameters and then sends the result back to the central controller. 


Suppose there are $N$ learners in the system, where $N\geq M$ and $M$ is the number of agents. We 
use a matrix $\bfC\in \mathbb{R}^{N\times M}$ to describe the agent-to-learner assignment. In particular, learner $j$ will update the parameters for agent $i$ if the $(j,i)$-th entry of the assignment matrix $\bfC$, denoted as $c_{j,i}$, satisfies $c_{j,i} \neq 0$. Let $\theta'_i$ be the parameters for agent $i$ after the update with gradient ascent along $\nabla_{\theta_i} V^{\bfpi}_i(\bfs)$. Each learner will send $y^{\prime}_{j}=\sum_{i=1}^Mc_{j,i}\theta^{\prime}_i$ back to the central controller.

In an uncoded distributed learning system, each learner updates the parameters for a single agent. In other words, different learners are responsible for different agents and only $M$ out of the $N$ learners are used. The assignment matrix 
$\bfC_{Uncoded} \in\mathbb{R}^{N\times M}$ has entries:
\begin{equation}
c_{j,i}  = \begin{cases}1 , & \text{if~} i= j,\\ 0, & \text{else}. \end{cases} \nonumber
\end{equation}


In this paper, we focus on synchronous learning systems. Therefore, the central controller does not update the agent policies until all updated parameters have been received. 
The learning efficiency is thus bounded by the slowest learner at each training iteration. Any node or link failure may affect the whole task. In the following subsection, we introduce a coded distributed learning framework to address this problem. 





\subsection{Coded Distributed Learning for MARL}
To enhance the resilience of a distributed learning system to uncertain stragglers, 
our idea is to introduce redundancies into the computation 
by assigning one or more agents to each learner. This is achieved by applying coding schemes to construct an assignment matrix $\bfC$ that satisfies $rank(\bfC)=M$ and has one or more non-zero entries in each row. 

In this coded learning framework, the central controller will be able to recover all updated parameters, denoted as $\bftheta^\prime=[\theta_1^{\prime T},\ldots,\theta_M^{\prime T}]^{T}$ with results received from only a subset of the learners. Particularly, let $\mathcal{I} = \{j~|~ y'_j \text{~is received} \}$ represent the set of learners whose results are received by the central controller by a certain time. Also let $\bfC_\mathcal{I} \in \mathbb{R}^{|\mathcal{I}|\times M}$ be a submatrix of $\bfC$ formed by the $j$-th rows of $\bfC$, $\forall j\in \mathcal{I}$. Then $\bftheta'$ can be recovered when $rank(\bfC_\mathcal{I})=M$ via:
\begin{equation}
\label{eq:decoding}
\bftheta'=(\bfC_\mathcal{I}^T\bfC_\mathcal{I})^{-1}\bfC_\mathcal{I}^T\bfy_{\calI}',
\end{equation}
where $\bfy_{\calI}'$ is the aggregate result derived by concatenating $y'_j$, $\forall j\in \mathcal{I}$. 
In the following subsection, we introduce four different coding schemes and explain how they can be used to construct a coded assignment matrix $\bfC$. 

\subsection{Coding Schemes}

\subsubsection{Replication-based Code} 
Using a replication-based coding scheme \cite{lee2017coded}, agents are assigned to the learners in a round-robin fashion and each agent is assigned to at least $\lfloor \frac{N}{M} \rfloor$ learners.  
The $(j,i)$-th entry of its assignment matrix $\bfC_{Replication}$ is then given by 
\begin{equation}
    c_{j,i} = \begin{cases} 1,& \text{if~} i = (j\text{~mod~}M) + M \mathbbm{1}_{(j\text{~mod~}M) = 0} \\
    0, &\text{else}\end{cases} \nonumber
\end{equation}
where $(j\text{~mod~}M)$ finds the remainder when $j$ is divided by $M$, and $\mathbbm{1}$ is the indicator function. 


\subsubsection{MDS Code}
An MDS code \cite{lacan2004systematic} specifies the assignment matrix such that any $M$ rows have full rank, by using, e.g., a Vandermonde matrix \cite{klinger1967vandermonde} as follows
\begin{equation}
\bfC_{MDS}=
 \begin{bmatrix}
1 & 1 & \cdots & 1 \\
\alpha_{1} & \alpha_{2} & \cdots & \alpha_{M} \\
\alpha_{1}^{2} & \alpha_{2}^{2} & \cdots & \alpha_{M}^{2} \\
\vdots & \vdots & \vdots & \vdots \\
\alpha_{1}^{N-1} & \alpha_{2}^{N-1} & \cdots & \alpha_{M}^{N-1}
\end{bmatrix} \nonumber
\end{equation}
where $\alpha_i$, $i \in [M]$, can be any non-zero real number. Note that all entries of $\bfC_{MDS}$ are non-zero, meaning that 
each learner needs to compute parameters for all the agents. 

\subsubsection{Random Sparse Code}
A random sparse code \cite{lee2017coded} enables a sparser assignment matrix $\bfC_{Random}$ with $(j,i)$-th entry randomly generated from a Gaussian distribution $\mathcal{N}(0,1)$ with probability $p_m$, i.e., $\mathbb{P}(c_{j,i} = \epsilon) = p_m$, where $\epsilon \sim \mathcal{N}(0,1)$. Otherwise, $c_{j,i} = 0$ with  $\mathbb{P}(c_{j,i} = 0) = 1- p_m$.  
Note that, 
by choosing an appropriate $p_m$, we can control the sparsity of the assignment matrix $\bfC_{Random}$.

\subsubsection{Regular LDPC Code}
The regular LDPC based coding scheme \cite{gabidulin2006rank} constructs an assignment matrix $\bfC_{LDPC}$ in three steps. The first step constructs a permutation matrix:
\begin{equation*}
    \mathbf{A}=\begin{bmatrix}
0 & 1 & 0 & \cdots & 0 \\
0 & 0 & 1 & \cdots & 0 \\
\cdots & \cdots & \cdots & \ddots & \cdots \\
0 & 0 & 0 & \cdots & 1 \\
1 & 0 & 0 & \cdots & 0
\end{bmatrix} \in \mathcal{F}_2^{w\times w}, 
\end{equation*}
where $\mathcal{F}_2$ represents the binary field and $w$ is a prime number that satisfies $\frac{N}{w} \in \mathbb{Z}^+$, where $\mathbb{Z}^+$ represents positive integers. The second step constructs a parity check matrix:
\begin{equation*}
   \mathbf{H} \!=\!\begin{bmatrix}
\mathbf{I}_{w} & \mathbf{I}_{w} & \mathbf{I}_{w} & \cdots & \mathbf{I}_{w} \\
\mathbf{I}_{w} & \mathbf{A} & \mathbf{A}^{w} & \cdots & \mathbf{A}^{w-1} \\
\cdots & \cdots & \cdots & \cdots & \cdots \\
\mathbf{I}_{w} & \mathbf{A}^{w-2} & \mathbf{A}^{2(w-2)} & \cdots & \mathbf{A}^{(w-2)(w-1)} \\
\mathbf{I}_{w} & \mathbf{A}^{w-1} & \mathbf{A}^{2(w-1)} & \cdots & \mathbf{A}^{(w-1)(w-1)}
\end{bmatrix} \!\in \!\mathcal{F}_2^{Y\times N}
\end{equation*}
where $Y$ satisfies $\frac{Y}{w} \in \mathbb{Z}^+$ and $Y \leq N$ and $\mathbf{I}_{w} \in \mathbb{R}^{w\times w}$ is an identity matrix. 
Finally, the assignment matrix is constructed by $\boldsymbol{C}_{LDPC}=\left[\mathbf{I}_{M}, \mathbf{P}\right]^T\ \in\mathcal{F}_2^{M\times N}$, where $\mathbf{P}$ is extracted from the parity check matrix with $\mathbf{H}=\left[-\mathbf{P}^{\top}, \mathbf{I}_{N-M}\right]$.
A positive feature of this coding scheme is that it can be quickly decoded by an iterative algorithm introduced in \cite{richardson2008modern}. The complexity of this algorithm is $\mathcal{O}\left(M\right)$, while the decoding complexity of other schemes is $\mathcal{O}\left(M^{3}\right)$. 

\section{Coded Distributed MADDPG}
\label{sec:coded_dmaddpg}

In this section, we apply the coded distributed learning framework 
to enhance the training efficiency of MADDPG \cite{lowe2017multi} and its resilience to stragglers. 
Unlike the general formulation of MARL described in Sec.~\ref{sec:problem}, MADDPG adopts a deterministic policy $\pi_i(s_i)$ for each agent $i \in [M]$, which only depends on the local state $s_i$. 
The value function $Q_i^{\bfpi}(\bfs,\bfa)$ of agent $i$ still depends on the joint state $\bfs$ and joint action $\bfa$. Moreover, for each agent $i$, four neural networks are used to approximate its policy $\pi_i(s_i; \theta_{p,i})$, value function $Q_i^{\bfpi}(\bfs,\bfa; \theta_{q,i})$, target policy $\hat{\pi}_i(s_i; \hat{\theta}_{p,i})$, and target value function $Q_i^{\hat{\bfpi}}(\bfs, \bfa; \hat{\theta}_{q,i})$, respectively.  $\hat{\bfpi}=(\hat{\pi}_1, \hat{\pi}_2,...,\hat{\pi}_M)$ denotes the concatenated target policy. Therefore, $\theta_i = [\theta_{p,i}, \theta_{q,i}, \hat{\theta}_{p,i}, \hat{\theta}_{q,i}]$. 
To optimize these parameters using the coded distributed learning framework, a central controller and multiple learners need to be implemented, whose interactions are described in detail as follows.

\subsection{Central Controller}
In each training iteration, the central controller generates policies for the agents based on the current parameters $\bftheta$. 
The agents then execute these policies  for several episodes and store the transitions $\crl{(\bfs, \bfa, \bfs', \bfr)}$ 
into a replay buffer $\calD$, where $\bfr = [r_1,r_2,\dots,r_M]$. The central controller samples a random mini-batch $\calB$ from the replay buffer $\calD$ and broadcasts the mini-batch $\calB$ and current parameters $\bftheta$ to the learners. After that, it waits for the learners to update the parameters and return back the results. 
As soon as the central controller receives sufficient results, it  recovers the updated parameters $\bftheta'$, 
sends acknowledgements to the learners and then proceeds to the next iteration. This procedure is summarized in Algorithm \ref{alg:codedMADDPG} (Line 1-15). 

\subsection{Learners}

When each learner $j$ receives parameters $\bftheta$ and  mini-batch $\calB$ from the central controller, it updates the parameters $\theta_i$ for each agent $i$ assigned to it, where $i \in [M], c_{j,i} \neq 0$.  
Specifically, the value function parameters $\theta_{q,i}$ are updated by minimizing the temporal-difference error:
\begin{equation}
\label{eq:maddpg_loss}
J(\theta_{q,i}) = \frac{1}{|\calB|} \!\sum_{(\bfs,\bfa,\bfs',r)\in \mathcal{B}} \!\! \prl{ L_i^{\hat{\bfpi}}(\bfs',r_i)\!-\!Q_{i}^{\bfpi}\prl{\bfs, \bfa ; \theta_{q,i}} }^{2}
\end{equation}
where $L_i^{\hat{\bfpi}}(\bfs',r_i)=r_i + \gamma Q_i^{\hat{\bfpi}}(\bfs',\hat{\bfpi}(\bfs'))$. The policy parameters $\theta_{p,i}$ are updated using gradient ascent, according to the policy gradient theorem \cite{RLBook}:
\begin{gather}
    \label{eq:maddpg_actor}
    \nabla_{\theta_{p, i}} J(\theta_{p, i}) \approx \frac{1}{|\calB|} \sum_{(\bfs,\bfa,\bfs',r)\in \mathcal{B}} \nabla_{\theta_{p, i}} \pi_i(s_i;\theta_{p, i}) \nabla_{a_i} Q_{i}^{\bfpi}\prl{\bfs, \bfa}.
\raisetag{2ex}
\end{gather}
The target policy and value function parameters are updated via Polyak averaging:
\begin{equation}
\begin{gathered}
\label{eq:target_update}
\hat{\theta}_{p,i} \leftarrow \tau\hat{\theta}_{p,i}+(1-\tau)\theta_{p,i}\\
\hat{\theta}_{q,i} \leftarrow \tau\hat{\theta}_{q,i} +(1-\tau)\theta_{q,i},
\end{gathered}
\end{equation}
where $\tau \!\in\! (0,1)$ is a hyperparameter. The procedure followed by the learners is summarized in Algorithm \ref{alg:codedMADDPG} (Line 16-26).

\begin{algorithm}[t] 
    \DontPrintSemicolon
    \tcp{\textbf{Central controller:}}
     Initialize $\bftheta$ for all agents\\
    \For{$k = 1:\text{max\_iteration}$}
    {
        \For{$\ell = 1:\text{max\_episode\_number}$}
        {
        \For{$t = 1:\text{max\_episode\_length}$}
        {
            For each agent $i$, select $a_{i,t}^\ell \!=\! \pi_i(s_{i,t}^\ell; \theta_{p,i})$.\\
            Execute joint action $\bfa_t^\ell$ and receive new state $\bfs_{t+1}'^{\ell}$ and reward $\bfr_{t+1}^\ell$.\\
            Store $\!(\bfs_t^\ell, \bfa_t^\ell, \bfr_{t+1}^\ell, \bfs_{t+1}'^\ell)\!$ in replay buffer $\!\mathcal{D}$.
        }
        }
        Sample a random minibatch $\mathcal{B}$ from  $\mathcal{D}$.\\
        Broadcast $\mathcal{B}$ and $\bftheta$ to the learners.\\
        $\bfy'\leftarrow[\ ]$\\
        
        \Do{$\bftheta'$ is not recoverable}
        {
        Listen to the channel and
        collect results $y'_j$ from the learners: $\bfy'\leftarrow[\bfy';y'_j], j \in [N]$
        }
        Send acknowledgements to learners.\\
        Recover $\bftheta'$ 
        using \eqref{eq:decoding} and let $\bftheta\leftarrow\bftheta'$
    }
    \tcp{\textbf{Learner $j$:}}
    \For{$k = 1:\text{max\_iteration}$}
    { 
    Listen to the channel. \\
    \If{receiving $\mathcal{B}$ and $\bftheta$ from central controller}
    {
            $y_j\leftarrow0$; $i \leftarrow 1$\\
            \While{$i \leq M$ and no acknowledgement received}{
            \If{$c_{j,i}\not = 0$}{
            Update $\theta_{p,i}$ using gradient ascent with gradient computed using  \eqref{eq:maddpg_actor}.\\
           Update $\theta_{q,i}$ by minimizing the loss in  \eqref{eq:maddpg_loss} with gradient descent.\\
            Update $\hat{\theta}_{p,i}$ and $\hat{\theta}_{p,i}$ using  \eqref{eq:target_update}.
            $y'_j \leftarrow y'_j + c_{j,i}\theta_i$}
            $i\leftarrow i+1$\\
            }
            Send $y'_j$ to the central controller.\\
            }
     }  
\caption{Coded Distributed MADDPG} \label{alg:codedMADDPG}
\end{algorithm}

\section{Experiments}
\label{sec:experiments}

We conduct a variety of experiments in the robotics environments proposed in \cite{lowe2017multi} to evaluate the performance of our coded distributed MADDPG algorithm.
We first describe the environments and then present the results.  

\subsection{Environments}
\begin{figure*}[t]
\centering
\vspace*{-0.4em}
	\subfigure[]{\includegraphics[width=0.21\textwidth]{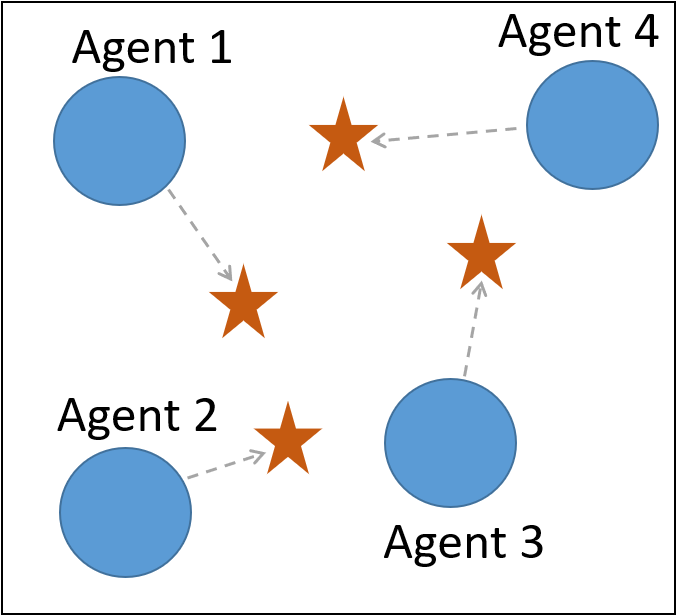}\label{fig:cooperative_navigation}
	}\vspace*{-0.2em}\hspace*{1.5em}
	\subfigure[]{\includegraphics[width=0.21\textwidth]{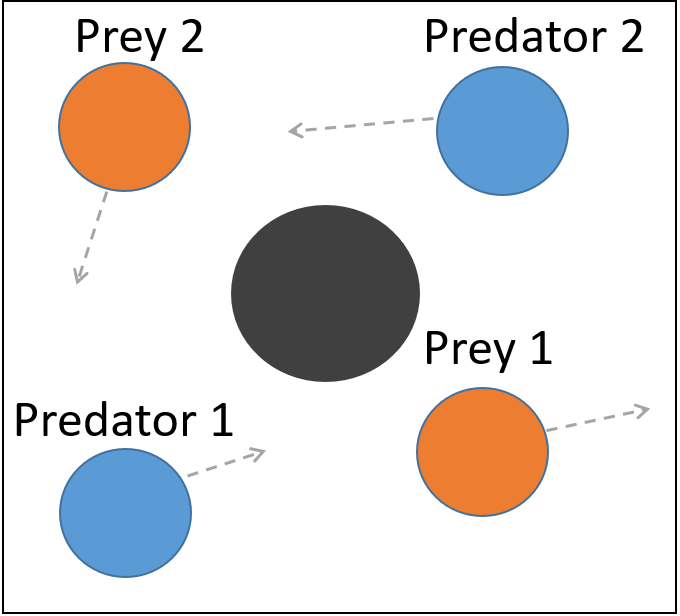}\label{fig:predator}
	}\vspace*{-0.2em}\hspace*{1.5em}
		\subfigure[]{\includegraphics[width=0.21\textwidth]{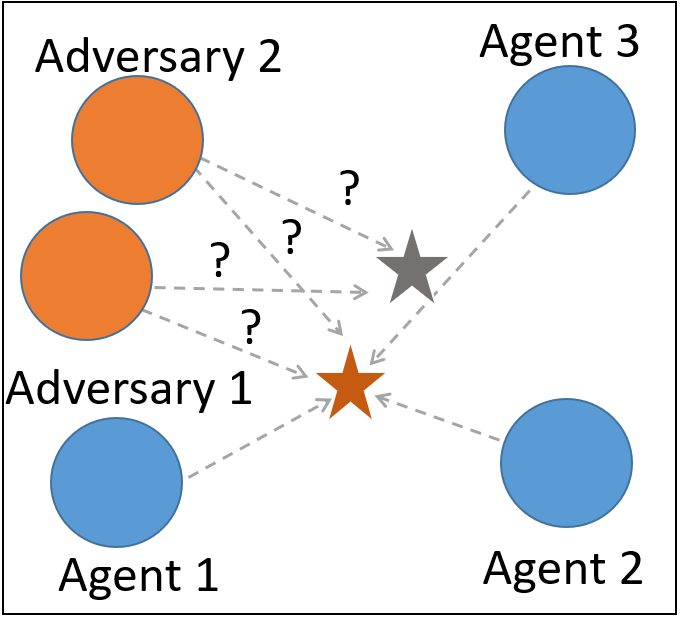}\label{fig:physical_deception}
	}\vspace*{-0.2em}\hspace*{1.5em}
	\subfigure[]{\includegraphics[width=0.21\textwidth]{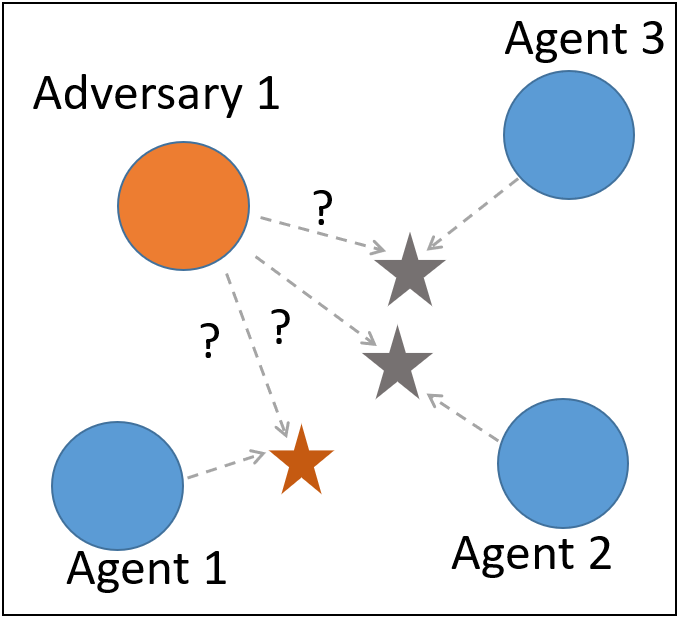}\label{fig:push}
	}\vspace*{-0.1em}
	\caption{Illustration of the experimental environment for (a) cooperative navigation, (b) predator-prey, (c) physical deception, and (d) keep away.}
	    \label{fig:environments}
\end{figure*}

The following four environments are considered. The agent interactions are either cooperative, competitive or mixed.  
\begin{itemize}
\item \textbf{Cooperative navigation:} In this environment, $M$ agents aim to cooperatively reach $M$ landmarks, while avoiding collisions with each other. The landmarks are not assigned to the agents explicitly. Therefore, the agents must learn to reach all landmarks by themselves. The agents are collectively rewarded based on the closeness of any agent to each landmark and are penalized if collisions happen (see Fig. \ref{fig:cooperative_navigation}).

\item \textbf {Predator-prey:} In this environment, $M-K$ good agents with restricted speed of motion aim to cooperatively chase $K$ faster adversary agents while avoiding static obstacles. When a good agent collides with an adversary agent, the good agents are rewarded while the adversary agents are penalized
(see Fig. \ref{fig:predator}).

\item \textbf {Physical deception:} In this environment, $M-1$ good agents aim to cooperatively reach a single known target among $L$ landmarks. One adversary agent also desires to reach the target landmark but does not know which one is the target and must infer it from the movements of the good agents. Therefore, the good agents must learn to spread out to cover all landmarks so as to confuse the adversary. The good agents are rewarded based on the minimum distance of any good agent to the target landmark and are penalized based on the distance of the adversary to the target. The adversary is rewarded based on its distance to the target landmark (see Fig. \ref{fig:physical_deception}).

\item \textbf {Keep away:} 
This environment is similar to the physical deception, but with $M-K$ good agents and $K$ adversary agents. Additionally, 
the adversaries can get in the way to prevent the good agents from reaching the target, and both the good and adversary agents are rewarded based on their distances to the target landmark (see Fig. \ref{fig:push}).

\end{itemize}

\begin{figure*}[t]
\centering
\vspace*{-0.4em}
\hspace*{-1.92em}
		\subfigure[]{\includegraphics[width=0.28\textwidth]{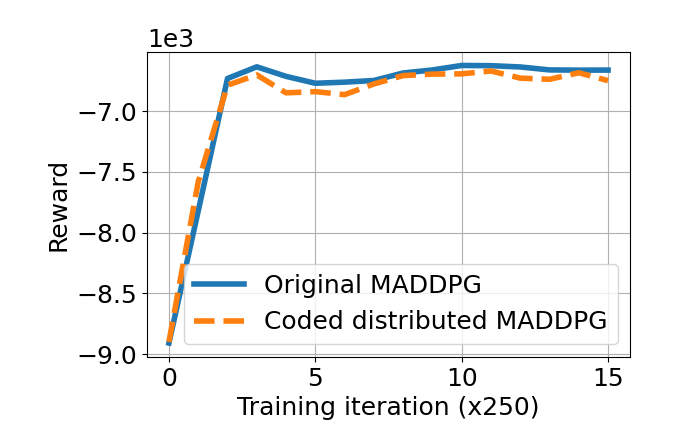}\label{fig:spread_reward}
	}\vspace*{-0.25em}\hspace*{-2.12em}
	\subfigure[]{\includegraphics[width=0.28\textwidth]{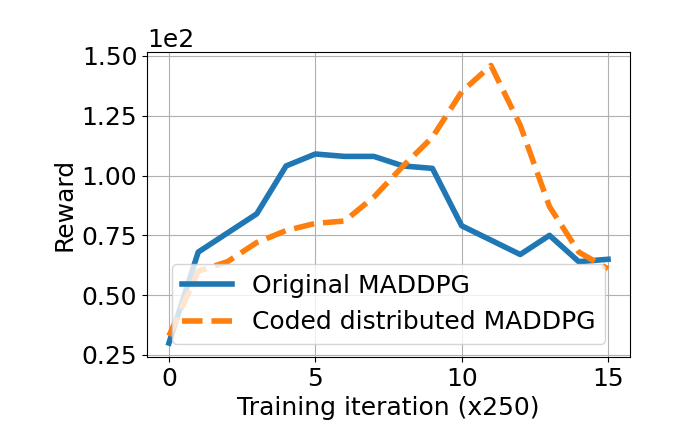}\label{fig:tag_reward}
	}\vspace*{-0.25em}\hspace*{-2.12em}
		\subfigure[]{\includegraphics[width=0.28\textwidth]{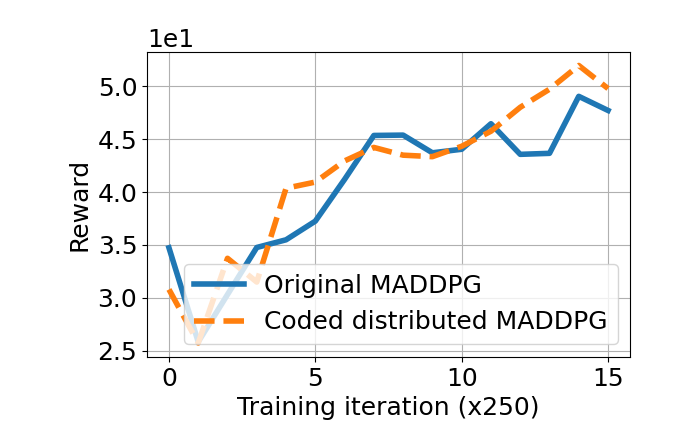}\label{fig:adversary_reward}
	}\vspace*{-0.25em}\hspace*{-2.12em}
		\subfigure[]{\includegraphics[width=0.28\textwidth]{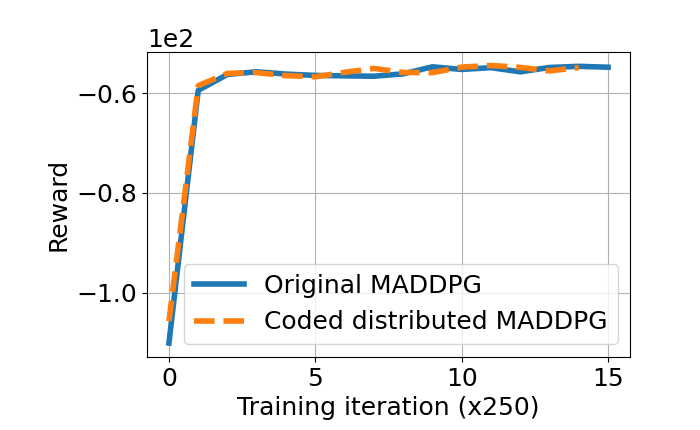}\label{fig:push_reward}
	}\vspace*{-0.1em}\hspace*{-1.92em}
	\caption{Average cumulative training reward on (a) cooperative navigation, (b) predator-prey, (c) physical deception, and (d) keep away.}
	    \label{fig:training_reward}
\end{figure*}

\subsection{Training Reward Comparison}
To demonstrate the effectiveness of the proposed coded distributed learning framework, we compare the coded distributed MADDPG with the original centralized MADDPG. The total number of agents is set to $M=8$. In the competitive environments, the number of adversary agents is set to $K=4$. 
The cumulative rewards of all agents averaged over 250 training iterations  for each environment is shown in Fig. \ref{fig:spread_reward}-\ref{fig:push_reward}. As we can see, in all environments, the coded distributed MADDPG is able to generate the same quality of policies as the original MADDPG and converges within the same number of iterations. 

\begin{figure*}
\centering
\vspace*{-0.4em}
	\subfigure[]{\includegraphics[width=0.265\textwidth]{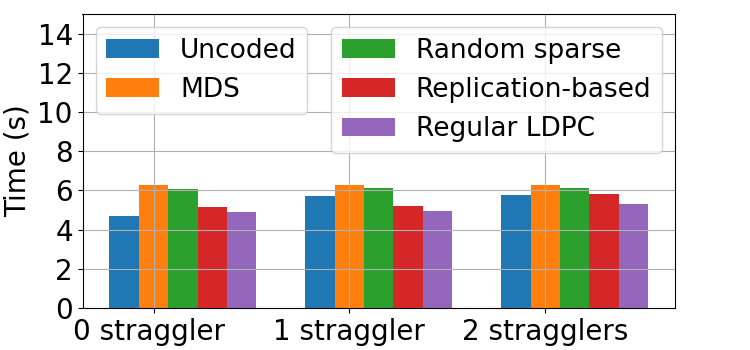}
	    \label{fig:time_exp2_8}
	}\vspace*{-0.25em}\hspace*{-1.2em}
	\subfigure[]{\includegraphics[width=0.265\textwidth]{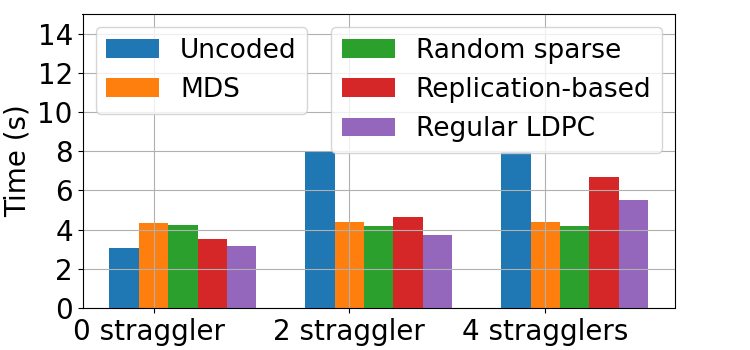}
	    \label{fig:time_exp1_8}
	}\vspace*{-0.25em}\hspace*{-1.2em}
	\subfigure[]{\includegraphics[width=0.265\textwidth]{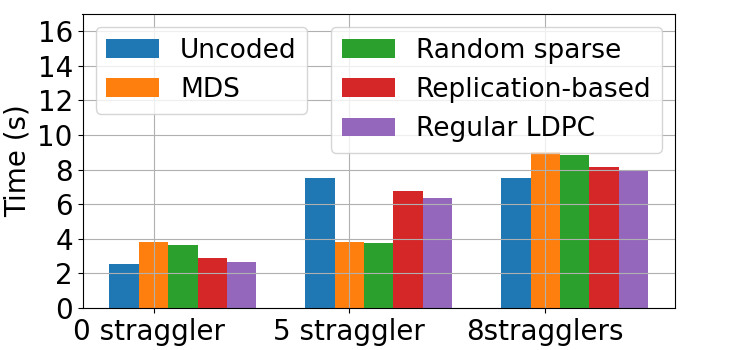}
	    \label{fig:time_exp3_8}
	}\hspace*{-1.2em}
		\subfigure[]{\includegraphics[width=0.265\textwidth]{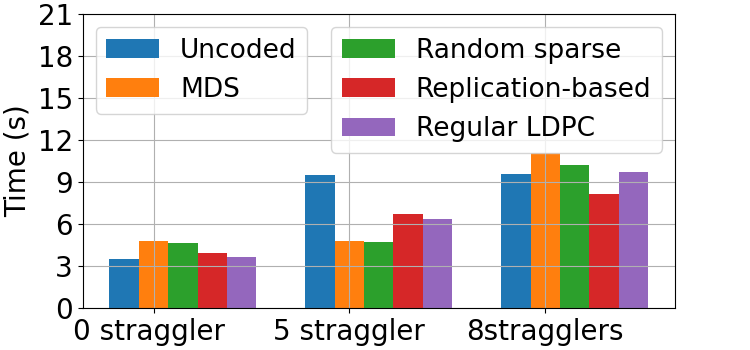}
	    \label{fig:time_exp4_8}
	}
	\vspace*{-0.25em}
	\caption{Average training time 
	on (a) cooperative navigation, (b) predator prey, (c) physical deception, and (d) keep away, when $M=8, N=15$.}
	    \label{fig:training_time_8}
\end{figure*}

\begin{figure*}
\centering
\vspace*{-0.4em}
	\subfigure[]{\includegraphics[width=0.265\textwidth]{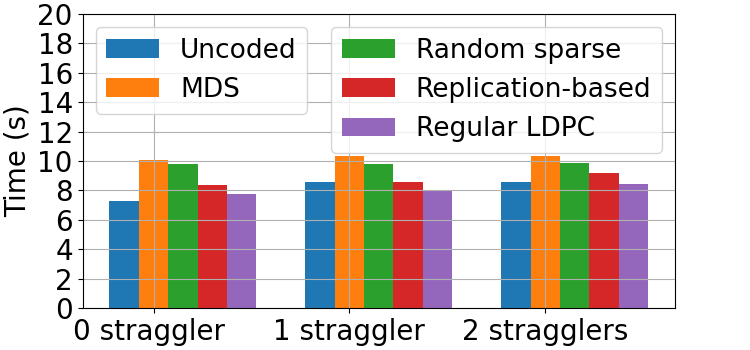}
	    \label{fig:time_exp2_10}
	}\vspace*{-0.25em}\hspace*{-1.2em}
	\subfigure[]{\includegraphics[width=0.265\textwidth]{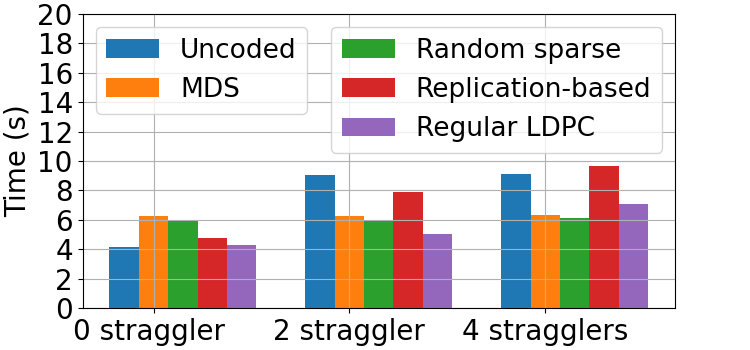}
	    \label{fig:time_exp1_10}
	}\vspace*{-0.25em}\hspace*{-1.2em}
	\subfigure[]{\includegraphics[width=0.265\textwidth]{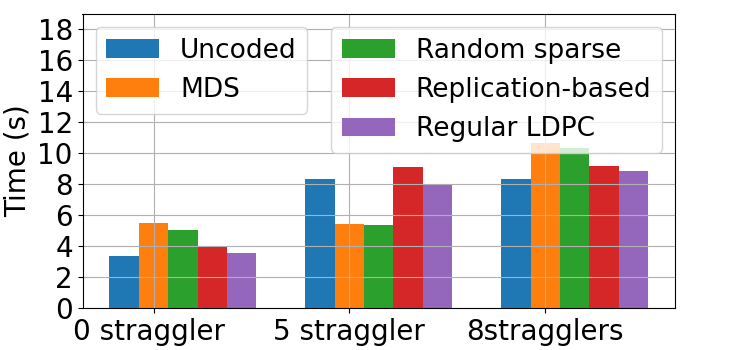}
	    \label{fig:time_exp3_10}
	}\vspace*{-0.25em}\hspace*{-1.2em}
		\subfigure[]{\includegraphics[width=0.265\textwidth]{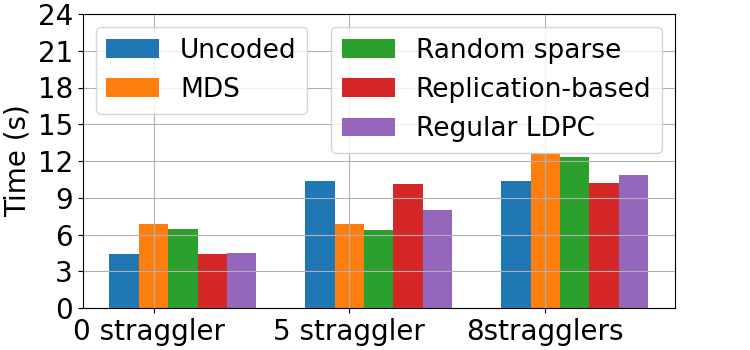}
	    \label{fig:time_exp4_10}
	}
	\vspace*{-0.1em}
	\caption{Average training time 
	on (a) cooperative navigation, (b) predator prey, (c) physical deception, and (d) keep away, when $M=10, N=15$.}
	    \label{fig:training_time_10}
\end{figure*}

\subsection{Training Time Comparison}
To evaluate the efficiency of the proposed coded distributed learning framework, we compare the training time of the coded distributed MADDPG with the uncoded distributed MADDPG. Different coding schemes, including the replication-based code, MDS code, random sparse code and regular LDPC are  evaluated.

To implement the distributed learning systems, we used Amazon EC2 and chose m5n.large \cite{m5nlarge} instances to implement the central controller and learners. To simulate uncertain stragglers, we randomly pick  
$k$ learners at each training iteration as stragglers, which delay returning the results for 
$t_s$ seconds. 
The specific experimental settings adopted for each environment are summarized as follows:
\begin{itemize}
\item \textbf{Cooperative navigation:} $k \in \{0, 1, 2\}$, $t_s=0.25s$.
\item \textbf{Predator prey:} $k\in \{0, 2, 4\}$, $t_s=1s$.
\item \textbf{Physical deception:} $k\in\{0, 5, 8\}$, $t_s=1s$.
\item \textbf{Keep away:} $k\in\{0, 5, 8\}$, $t_s=1.5s$.
\end{itemize}
Besides varying the value of $k$ in each environment, we also vary the total number of agents $M$. Results obtained when $M=8$ and $M=10$ are shown 
in Fig. \ref{fig:training_time_8} and Fig. \ref{fig:training_time_10}, respectively. Each value represents the training time of an algorithm averaged over 5 iterations, with the total number of iterations set to 50. In all experiments, the total number of learners is set to $N=15$, and the parameter $p_m$ in the random sparse code is set to $0.8$.
 
Analyzing the two figures, we can make the following observations. First, when there are no stragglers, the uncoded distributed MADDPG  achieves the best performance in all experiments. This demonstrates the good performance of the traditional uncoded distributed learning framework in ideal scenarios without stragglers.

When stragglers are present, we can observe that the performance of the uncoded scheme degrades significantly and achieves worse performance than most coding schemes. Furthermore, its performance remains stable as the number of stragglers increases. This is because each straggler is delayed by the same amount of time, $t_s$, such that each iteration is always delayed by $t_s$ no matter how many stragglers are present.   
In contrast, the coding schemes are generally more robust to stragglers, except when 
the number of stragglers exceeds the limit that the coding schemes can tolerate. 

Next, we analyze the performance of different coding schemes. We can observe that the MDS code (orange bars) is very robust to stragglers when the number of stragglers $k$ does not exceed the maximum tolerable number $N-M$. However, when $k> N-M$, the MDS performance degrades significantly, as shown in Fig. \ref{fig:time_exp3_8}-\ref{fig:time_exp4_8} and Fig.  \ref{fig:time_exp3_10}-\ref{fig:time_exp4_10}. Furthermore, we can observe that when $k<N-M$ and when the straggler effect, indicated by the amount of delay $t_s$ introduced by the stragglers, is relatively large, MDS outperforms the uncoded scheme, as well as the replication-based and regular LDPC codes, as shown in Fig. \ref{fig:time_exp1_8}-\ref{fig:time_exp4_8} and Fig. \ref{fig:time_exp1_10}-\ref{fig:time_exp4_10}. However, when $t_s$ is relatively small, the MDS code has the worst performance, as shown in Fig. \ref{fig:time_exp2_8} and \ref{fig:time_exp2_10}, due to the high computational redundancies it introduces through the dense assignment matrix.  

The random sparse code (green bars) shows a similar performance to the MDS code. This is because when its parameter $p_m$ takes a large value ($p_m = 0.8$ in our experiments), the assignment matrix generated by this type of code has a similar density as the one generated by the MDS code. 

Finally, both the replication-based code (red bars) and regular LDPC code (purple bars) are more affected by an increase in the stragglers, as their assignment matrices are sparser. However, they achieve better performance than MDS and random sparse codes when the straggler effect is small, 
as shown in Fig. \ref{fig:time_exp2_8} and \ref{fig:time_exp2_10}, and when there are many stragglers, 
as shown in Fig. \ref{fig:time_exp3_8}-\ref{fig:time_exp4_8} and Fig. \ref{fig:time_exp3_10}-\ref{fig:time_exp4_10}. 

\section{Conclusion}
\label{sec:conclusion}
This paper introduces a coded distributed learning framework for MARL, which 
improves the training efficiency of policy gradient algorithms in the presence of stragglers while not degrading the accuracy. We applied the proposed framework to a coded distributed version of MADDPG, a start-of-the-art MARL algorithm. Simulations on several multi-robot problems, including cooperative navigation, predator-prey, physical deception and keep away tasks, demonstrate the high training efficiency of the coded distributed MADDPG, compared with the traditional uncoded distributed learning approach. The results also show that the coded distributed MADDPG generates policies of the same quality as the original centralized MADDPG and converges within the same number of iterations. Furthermore, we investigated different     
coding schemes including replication-based, MDS, random sparse, and regular LDPC codes. 
Simulation results show that the MDS and random sparse codes can tolerate more stragglers but introduce larger computation overhead. Additionally, the replication-based and regular LDPC codes produce less overhead but are more susceptible to stragglers.

\balance
\begin{small}
	\bibliographystyle{IEEEtran}
	\bibliography{main}
\end{small}

\end{document}